# Patterns of Language - A Population Model for Language Structure

Robert John Freeman
rjfreeman@usa.net

*A key problem in the description of language structure is to explain its contradictory properties of specificity and generality, the contrasting poles of formulaic prescription and generative productivity. I argue that this is possible if we accept analogy and similarity as the basic mechanisms of structural definition. As a specific example I discuss how it would be possible to use analogy to define a generative model of syntactic structure.*

## 1.0 Introduction

Several authors have emphasized the importance of fixed formulaic structure for the full description of English syntax, e.g. Nattinger (1980), Pawley and Syder (1983) and others. Weinert (1995) provides a good review of current theory. A major theme of her article is the need for a coherent model of language structure to explain its dual aspects of particularity and generality. Most current discussions hypothesize that there are two competing mechanisms: a fixed formulaic memory, and an analytic, rule based, generator. In this paper I seek to propose a definition of language structure which can explain both the particularity and generality of language as different aspects of a single process.

## 2.0 Patterns of Language

In "Collocational Grammar" (Freeman, 1996) I discussed the potential of analyzing language in terms of groups of tokens, rather than logical relationships between individual tokens. Essentially I was arguing that we should think about language structures most fundamentally as patterns. The most important tool for the efficient representation of structure under that framework was a concept of similarity between patterns. Apart from discussing the most efficient representation, I showed that it was possible to define a concept of lexical class, in terms of similarities between collocational contexts, which accurately modeled our traditional perceptions of lexical syntactic class. I did not, however, discuss explicitly how one might model a generative syntax in terms of collocational patterns alone.

## 3.0 Generation of Syntactic Patterns

Traditionally it has been assumed that, as we perceive syntactic relationships to be logical, our definitions of syntactic structure should be logical, and new collocational patterns should be generated by a process of logical modifications of existing patterns. This is the kind of generality Krashen (1978), for example, tests for when he explores the relationship between formulaic language and language acquisition. (Krashen finds that while all studies of acquisition indicate an element of formulaic language use, subjects do not show any tendency to modify logically the structures they have learned, "See you." to "I can see you." for instance, he therefore concludes that formulaic language is incidental to productive acquisition.) If we concentrate on form, however, rather than the formal description of form, it is much more natural to generalize structure in terms of similarities.

### 3.1 Populations
An example of a classification of structure based on form is that of a human population. Populations are made up of large numbers of individuals, each one logically distinct. They do, however, have a certain regular structure. Each individual is different, but we can group the individuals into families, communities, groups and sub-populations of various types. We do this through a concept of similarity.

### 3.2 Grammatical Acceptability
In the same way that we can define membership of a human population in terms of overall similarities between any individual and the members making up that population, if we think about syntactic patterns purely as form we can classify syntactic structures in terms of the degree of similarity between a proposed pattern and all the accepted patterns of a language.

What role could we find for this classification in a formal description of language? Let us use it to define the acceptability of a given syntactic pattern, i.e. we define the acceptability of a syntactic pattern as the similarity between it and the sum total of existing, known, patterns of the language. Each of the structures in the population of established patterns is unique, but to an extent they are all the same. Naturally if a proposed structure is exactly the same as any existing pattern it will be judged acceptable. If it is just a little different it may still be regarded as belonging to one family or another. According to the degree of difference between a proposed pattern and the existing patterns it can be judged acceptable in more and more general terms. If we define the acceptability of a given combination of words as the extent to which it is similar to sub-sets at various levels, to the population as a whole, but also to each of the individual members which make it up, then we have a definition for the acceptable structures of language with just the properties of infinite particularity and infinite generality that we find in natural language.

## 4.0 A Model for Grammatical Acceptability
As a working definition let us specify that a new syntactic combination is acceptable if all or most of its components can be found in the same order in other "individuals" of a given syntactic "population" grouping, with the level of acceptability given by the level of generality of that grouping. Say we have an expression:

- "Bar the door"

It is reasonable to assume there will be a "family" of collocations common to "door" and, say, "gate" which will include the following examples:

- "Bar the gate"
- "Shut the gate"
- "Shut the door"

and many others. According to our working definition the new expression will be acceptable because "bar" occurs in the context of the common family of "gate" and "door". Other collocational combinations will be blocked because they do not. Any word, however, will have a level of collocational appropriateness according to the level of generalization, the "family" or "sub-population" of form at which they were adjudged similar. If we consider the collocates with "hatch", "window", "screen", "curtain", in ever widening circles, then the collocational environment, or sub-population of form, would become ever more diverse. Collocations of "curtain" would often seem distinctly odd with "gate" or "door", but the process would be gradual, and even "*Draw the door" would be more acceptable than "*Zebra the door", for instance.

In the other direction any peculiarities of one word with respect to another would also have a natural representation. "Jump the gate" could be recorded as a particular expression in contrast to given uses of "door", and could have specific meaning associated with it as a unit in addition to productive intuitions derived from the common contexts of "gate" and any other word.

### 4.1 Representation Spaces of Form
The most efficient set of formulas or patterns in such a definition could be considered to be "factors" of a form-based representation of language structure. This is the "most efficient representation of structure" I discussed in "Collocational Grammar". (The idea of factorial representations of structure is a common one, even in natural language applications, though it has largely been limited to classifying entire documents, as in the "vector model" of content for document retrieval (e.g. Deerwester et al., 1990), and notably by Douglas Biber for the study of discourse structure (e.g. Biber, 1991). What is needed is the recognition that a given classification is based on similarity and the advantages of factorial representation follow naturally, the only difference is that historically this has not been considered to be the case for syntax.) To fully represent the syntactic restrictions of a language, in general, all the factors would be necessary. As more and more are removed the specification of form would become more and more general until only the most general aspects were left.

Consider the kind of collocational error which occurs in the language of learners of English. Howarth (1996) gives the example:

- *concerning proposals done by historians

which he characterizes as interference between two more natural native speaker forms: "proposals made" and "surveys done." Imagine a structural "representation" for this corner of English syntax made up of the four sentences:

- concerning proposals forwarded by historians (1)
- concerning things made by historians (2)
- concerning proposals made by historians (3)
- concerning surveys done by historians (4)

Then, by definition, "proposals made" is an acceptable sentence. If, however, we delete (3) from our representation of structure then the patterns:

- concerning proposals made by historians
- *concerning proposals done by historians

would be equally close to (1) and (2), and (1) and (4), respectively. There would be nothing to distinguish between them. As we continued to delete patterns selection would become worse and worse and our grammar would gradually deteriorate, with exactly the kind of graceful degradation that is such a characteristic feature of language "error" and the limited production skills of language learners.

## 5.0 Formal Grammar

My intention in this paper is simply to support the assertion that it is possible to specify language structure in terms of form alone, but it is a little hard to see how such a definition would relate to our normal concepts of logical structure and the objectives of syntactic analysis. It is worth justifying its utility and relating it in some way to more traditional perspectives.

In a way the distinction between our traditional models of structure and an analogical one is like the distinction between competence and performance traditionally made in generative linguistics, it certainly provides an explanation for the extent of the gap between the two in the usual formulation. Structure in an analogical model is *specified* (to the extent that it does not indicate the implementation of a meaningful choice on the behalf of a speaker) in terms of similarity, but each level of specification (or at least the population norms of each level of specification) will have a most efficient *description* in terms of a logical grammar (as will the elements of any random string). Description is competence, specification is performance (though, of course, we can only ever predict syntactically aspects of performance which are not subject to free choice, such as the statistical norms of corpora). It is worth noting that the earliest generative grammars never claimed to be any more than a most efficient description of possible language structures, the idea that the logical rules of generative grammars might provide a model for cognitive mechanisms of language processing was only ever a hypothesis. It might have been assumed by many, but it was not, I think, ever claimed to have been proven (e.g. Bever, 1970). In fact, then, defining structure in terms of analogy contradicts nothing in the existing theory of formal grammatical description; what is challenged is the correspondence theory of psycholinguistics, which asserts that the processes of generative grammar have exact psychological analogues which are important in the specification of language structure. What has gone in an analogical model is a physical realization of logical production, the actual structures are produced by analogical processes, the logical structure is an abstraction of common properties.

### 5.1 Logical Description.

Just as I think analogy will prove indispensable to model the processes of language, logical grammars will probably prove indispensable for their description. The only proviso would be that because of the fundamentally analogical definition we must recognize that a logical structure exists only for the population norms, or average forms.

Each member of a population has an intrinsic, logical, structure, and it would be possible to characterize a "typical" member of any population or sub-population in terms of this. Indeed, this is the only reasonable way it can be characterized. It is important to realize, though, that it is the members of a given population which specify the structure to be taken as typical and which thus play the fundamental role of definition, we can characterize populations logically, but we cannot specify the individuals which make them up purely in terms of this logical structure, there are too many variations of detail. This is the difference between a logical system and a pattern-based, or analogical, system. In a logical system the logical structure specifies the patterns. In a pattern based system the patterns specify the logical structure.

For example, consider a logical structure, like a clockwork mechanism, and an analogical structure, like the population of all businessmen. A clockwork is uniquely specified by its logical structure, all examples of a given clockwork will have exactly the same structure. A population, like the set of all businessmen, however, will have certain general features, a tendency to wear white shirts and neck ties perhaps, but the set of all businessmen will not be in any way specified by this logical structure. A characterization in terms of white shirts and neck ties will capture the general nature but not all of the structure of the sartorial class of businessmen.

On the other hand, just as we cannot define the clothing of all business men in a concise logical manner, it is impossible to describe it (as opposed to defining it) in a strictly analogical way. In answer to the question "how do businessmen dress" the answer, "like the set of all businessmen" is of little use. (At best we could find a typical example, a "factorial" representation, but even that does not tell us anything *about* the typical example, it only shows us what it is.) Similarly for language, while I claim we will need to define structure fundamentally in terms of populations of form, and while this insight will have very profound implications for how we model language performance automatically, or attempt to teach or learn a language, for description (in the sense that "the subject precedes the verb" is a description) we must rely on logical relationships within typical examples. In short, language structure should be defined analogically, but perceived logically.